\pdfoutput=1

\documentclass[11pt]{article}
\usepackage{arabtex}
\usepackage{utf8}
\setcode{utf8}
\usepackage{booktabs}
\usepackage{multirow}
\usepackage[russian, english]{babel}

\usepackage[final]{mtsummit25}
\usepackage{copyright}

\usepackage{times}
\usepackage{latexsym}
\usepackage{fancyvrb}
\usepackage{csquotes}
\usepackage[T1]{fontenc}

\usepackage[utf8]{inputenc}

\usepackage{enumitem}

\usepackage{microtype}

\usepackage{inconsolata}

\usepackage{graphicx}
\usepackage{listings}

\lstdefinelanguage{JSON}{
    basicstyle=\ttfamily\small,
    showstringspaces=false,
    breaklines=true,
    stringstyle=\color{blue},
    keywordstyle=\color{magenta},
    morestring=[b]",
    literate=
      *{:}{{\textcolor{orange}{:}}}1
       {,}{{\textcolor{orange}{,}}}1
       {\{}{{\textcolor{orange}{\{}}}1
       {\}}{{\textcolor{orange}{\}}}}1
       {[}{{\textcolor{orange}{[}}}1
       {]}{{\textcolor{orange}{]}}}1,
}

\usepackage{comments}
\usepackage{comment}

\newauthor[Sh]{Simon}{yellow}  
\newauthor[Pb]{Pierrette}{red}     
\newauthor[Pa]{Perla}{green}  
\usepackage{tipa}
\title{Arabizi {\it vs} LLMs: Can the Genie Understand the Language of Aladdin?}

\author{Perla Al Almaoui \\
  Faculté de traduction et d'interprétation \\
  Université de Genève \\
  \texttt{almaoui.perla@outlook.com} \\\And
  Pierrette Bouillon \\
  Faculté de traduction et d'interprétation \\
  Université de Genève \\
  \texttt{pierrette.bouillon@unige.ch} \\ \AND
  Simon Hengchen \\
  Faculté de traduction et d'interprétation \& iguanodon.ai \\
  Université de Genève \\
  \texttt{simon.hengchen@unige.ch} \\
  }

\begin{document}

\maketitle
\begin{abstract}
In this era of rapid technological advancements, communication continues to evolve as new linguistic phenomena emerge. Among these is Arabizi, a hybrid form of Arabic that incorporates Latin characters and numbers to represent the spoken dialects of Arab communities. Arabizi is widely used on social media and allows people to communicate in an informal and dynamic way, but it poses significant challenges for machine translation due to its lack of formal structure and deeply embedded cultural nuances.
This case study arises from a growing need to translate Arabizi for gisting purposes. It evaluates the capacity of different LLMs to decode and translate Arabizi, focusing on multiple Arabic dialects that have rarely been studied up until now. Using a combination of human evaluators and automatic metrics, this research project investigates the models’ performance in translating Arabizi into both Modern Standard Arabic and English. Key questions explored include which dialects are translated most effectively and whether translations into English surpass those into Arabic.\end{abstract}

\section{Introduction}
Although there are approximately 420 million Arabic speakers worldwide, an intriguing linguistic paradox emerges: Modern Standard Arabic (MSA), the standardized form of the language, is the mother tongue of none. Instead, Arabs communicate through their regional dialects, which are vibrant linguistic hybrids influenced by Arabic and the historical languages of each region. These dialects have been honed by geographic, cultural, and historical factors, and can vary significantly even within a single country, resulting in a mosaic of over 60 distinct varieties. 
Arabizi (a fusion of "Arabic" and Englizi, the Arabic word for English) is an informal, non-standard writing system that emerged in the 1990s when Arabic keyboards were not widely available. It uses Latin characters and numbers, combining both transliteration and transcription mappings. Primarily used in online communication—such as short messages and comments on social media—Arabizi varies significantly across dialects and even within the same dialect \cite{Harrat}. 
For instance, the transcription of the following sentence in MSA \<أريد أن أكلّمك بموضوع> (I want to talk to you about something) in Arabizi can be "badde e7kik bi mawdu3" in Levantine Arabic, or "rani hab nahdar m3ak f wahd sujet" in Algerian Arabic. 



The idea of romanizing the Arabic language is not a new concept, as there have already been several attempts to do so over the last century. However, these efforts largely failed, as they were perceived as colonialist initiatives aimed at suppressing cultural and religious identity. More recently, the International Organization for Standardization (ISO) introduced two norms, ISO 233 in 1984 and ISO 233-2 in 1993, to standardize the romanization of Arabic. These standards aimed to facilitate the international exchange of information. Nevertheless, their adoption remained limited due to their impracticality, with usage restricted primarily to official contexts \cite{almaoui2024chatgpt}.

Conversely, Arabizi has become the dominant written form of communication among Arabic speakers in informal settings. Its rise reflects a crucial sociolinguistic reality: while MSA remains the language reserved for academic, religious, and formal settings, it is often perceived as inaccessible or overly formal for daily use. Arabizi, by contrast, offers a dynamic and flexible medium for self-expression that aligns with the fluidity of Arabic dialects \cite{allehaiby2013arabizi,yushmanov1961structure}.

Despite its widespread use across digital platforms, and the recent focus on informal language and low-resource languages, Arabizi remains an unexplored area in natural language processing (NLP). It poses particular challenges due to its colloquial nature, variation across dialects and lack of standardization, as well as the scarcity of digital resources. In NLP, research on Arabizi has mainly focused on transliteration into Arabic (deromanization) at the character or word level, using different approaches \cite{guellil, shazal}, and on the creation of a parallel annotated corpus of SMS and chat messages written in Arabizi and their corresponding Arabic script transliterations by \cite{bies2014}.
Some studies explore interlinguistic machine translation (MT) techniques to and from Arabizi, 
employing various architectures and pipelines, mainly between English and Egyptian dialects (see \citet{Harrat} for a summary up until 2017). While some recent datasets for low resource language translation include romanization, they are not specifically focused on Arabizi. Flores benchmark \cite{Flores}, for example, is limited to the romanized transcription of MSA or Arabic dialects in Arabic scripts. The TerjamaBench dataset
\cite{atlasia2024terjamabench}
is an exception and includes entries in Darija, the Moroccan Arabic dialect, written in both Latin alphabet (Arabizi) and Arabic script, and their corresponding English translations.

Since there is a growing need to translate Arabizi into
resource-rich languages on social media and other
digital platforms, we conducted a case study to evaluate the feasibility of using large language models (LLMs) for out-of-the-box machine translation.
The project began when the language technology company iguanodon.ai received a request from a client who wanted to know if short Arabizi texts could be translated for gisting purposes. The study involves a collaboration between the start-up and a professional
translator with previous experience in Arabizi.
Our contribution includes \mbox{AladdinBench \includegraphics[height=1em]{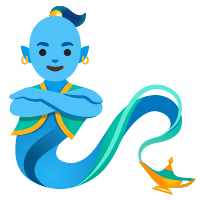}}, an authentic dataset in Arabizi for three dialects publicly available on huggingface\footnote{\url{https://huggingface.co/datasets/palmaoui/AladdinBench}} and a comparative evaluation of five LLMs using different prompting strategies.

To our knowledge, this is the first study that has explored the direct translation from different dialects in Arabizi to MSA or English without prior deromanization.

The rest of this paper is organized as follows. We describe the data production methodology and resulting dataset in Section 2, followed by the experimental  setup in Section 3 and the results in Section 4. Finally, discussion of results and the limitations of this study are presented in Section 5 and Section 6.

\section{Data Collection and Dataset}

\subsection{Dialects}
We decided to focus on the translation of three Arabic dialects from three distinct countries—Lebanon, Egypt, and Algeria—into two target languages: MSA, a less-resourced language, and English. These three countries were selected because their dialects represent distinct linguistic varieties. The Lebanese dialect aligns with the Levantine group and the Algerian dialect with the Maghrebi group, while the Egyptian dialect is exceptionally prominent due to its widespread use and cultural influence \cite{ayoubi2022arabic}.


The Lebanese dialect reflects a rich history and various cultural influences. Ancient languages such as Aramaic and Syriac, once dominant in northern Lebanon, had a notable impact on the dialect, particularly when it comes to phonological features like the use of silent vowels. Other regions, closer to major coastal cities, feature dialects more aligned with Classical Arabic, with fewer phonological deviations. Lebanon's Ottoman past also shaped its linguistic landscape, with Turkish loanwords becoming integral to Lebanese lexicon after four centuries of Ottoman rule \cite{Iskandar2022, almaoui2024chatgpt}.

Egyptian Arabic evolved through layers of historical migrations, demographic shifts, and ancient linguistic roots. It was heavily influenced by Coptic, the language of ancient Egypt, and later by Arabic after the Islamic conquest in the 7th century. Over time, Egyptian Arabic absorbed linguistic elements from Greek, Turkish, Italian, French, and English during various periods of occupation and cultural exchange. Regional variations within Egypt further enrich its linguistic diversity: northern regions, including the Delta and Cairo, feature subtle dialectal differences, while Upper Egypt's Sa'idi Arabic retains more conservative features. Additionally, Bedouin communities in the Western Desert speak Arabic varieties that are distinct from urban Egyptian Arabic \cite{souag, magidow2021old}.

Algerian Arabic is a product of extensive historical and cultural interactions. Indigenous Berber languages, particularly Tamazight, form its linguistic foundation, while successive occupations introduced other influences. The Roman era brought Latin, especially in administration and religion; the influence of this language was then further reinforced by Christian scholars such as Saint Augustine. The Arab conquest in the 7th century made Arabic the language of faith and the elite. Tamazight continued to be used in day-to-day life. Subsequent occupations by the Spanish, the Ottomans, and the French contributed lexical and structural elements to the dialect. French, in particular, had a profound impact during colonial rule, shaping Algeria’s modern plurilingual society.
Algerian Arabic is marked by significant regional variation. Western regions display a strong Spanish influence, while central areas, including Algiers, are heavily influenced by French. Eastern regions, such as Constantine, retain more Classical Arabic features. Southern regions, including the Sahara, exhibit notable Berber linguistic characteristics, reflecting the enduring presence of Berber-speaking populations \cite{saadanehabash, Chami2009}.

\subsection{Participants}
Thirty-one participants were recruited for the study through LinkedIn and targeted recruitment messages, with at least four participants per sub-dialect to ensure balanced representation. All participants were native Arabic speakers who represent the specific regional varieties outlined in the previous section. Lebanese participants were selected from both southern and northern regions of Lebanon. Similarly, Algerian participants were drawn from Algiers, the capital, and Constantine to reflect distinct linguistic traits within the country. For Egypt, participants were recruited from Cairo in the north and Luxor in the south.

Participants were asked to share WhatsApp conversations they had engaged in with peers of a similar age group (20–35 years) and from the same regions as them. These conversations revolved around a range of everyday topics, in order to reflect natural and spontaneous interactions. The focus on this age demographic provided a degree of consistency in communication styles, as participants shared a common digital literacy and texting culture.

\begin{table*}[h]
    \centering
    \small
    \renewcommand{\arraystretch}{1.2} 
    \setlength{\tabcolsep}{4pt} 
    \begin{tabular}{l l c c c c c c c c c}
        \toprule
        \textbf{Country} & \textbf{Region} & \textbf{Number of} & \textbf{Number of} & \textbf{Total} & \textbf{Tokens per} & \textbf{English} & \textbf{French} & \textbf{Mixed} & \textbf{\% code-switching} \\
        & & \textbf{segments} & \textbf{tokens} & \textbf{tokens} & \textbf{segment} & \textbf{words} & \textbf{words} & \textbf{words} & \textbf{in corpus} \\
        \toprule
        \multirow{2}{*}{Lebanon} & North  & 127 & 508 & \multirow{2}{*}{1075} & \multirow{2}{*}{3.1} & 55 & 12 & 0 & 13.19\% \\
                                 & South  & 141 & 567 &  &  & 33 & 11 & 0 & 7.76\% \\  
        \midrule
        \multirow{2}{*}{Egypt} & Cairo & 117 & 601 & \multirow{2}{*}{1159} & \multirow{2}{*}{8.1} & 28 & 0 & 0 & 4.65\% \\
                               & Luxor & 42 & 558 &  &  & 3 & 0 & 0 & 0.5\% \\  
        \midrule
        \multirow{2}{*}{Algeria} & Algiers & 145 & 639 & \multirow{2}{*}{1164} & \multirow{2}{*}{3.8} & 59 & 3 & 8 & 10.95\% \\
                                 & Constantine & 99 & 525 &  &  & 52 & 1 & 5 & 11\% \\  
        \bottomrule
    \end{tabular}
    \caption{Summary of segment, token, and foreign word counts by region}
    \label{tab:merged_summary}
    \vspace{-1em}
\end{table*}

All participants, including both recruits and their peers, signed consent forms explicitly detailing the use and processing of their data in accordance with Swiss law. After the corpus was collected, it was manually anonymized to ensure privacy, and all real names were removed and substituted with fictitious ones where necessary. Subsequently, a professional Arabic-speaking translator translated the corpus into MSA and English, with these translations serving as reference texts for automatic metrics. The translation into Arabic represents an intralingual transformation from a dialectal and informal variety of Arabic to a formal and standardized form. 



Table~\ref{tab:merged_summary} presents the collected corpora, including the number of segments and tokens, the average number of tokens per sentence and the percentage of foreign and mixed words (code-mixing). Mixed words are created by combining roots from one language with prefixes, suffixes, or morphological patterns from another language, reflecting linguistic creativity and contact-induced change.

\section{Experimental Setup}
We carried out a systematic evaluation of translation quality using an automated protocol. 
For each dialect, we created a combination of parameters defined as follows:
\vspace{-0.5em}
\begin{itemize}
    \small
    \setlength{\itemsep}{0pt}  
    \setlength{\parskip}{0pt}
    \item target language $\in$ [EN, MSA]
    \item prompt\_language $\in$ [EN, MSA]
    \item prompt\_strategy $\in$ [no-shot, one-shot, two-shot]
    \item prompt\_variation $\in$ [Lebanon, Egypt, Algeria]\footnote{More on this in Subsection~\ref{subsec:prompts}.}
    \item model $\in$ [GPT-4o, Llama 3, Claude, Gemini, Gemma, Mistral, Jais].
    \vspace{-0.5em}
\end{itemize}

All models were prompted with a temperature of $0.5$.
A discussion of the chosen models and prompts is available in \ref{subsec:models} and in \ref{subsec:prompts}. Evaluation metrics are presented in \ref{subsec:eval}.
Our code is available.\footnote{\url{https://github.com/iguanodon-ai/ArabizivsLLMs}}

\subsection{Models}
\label{subsec:models}
The models used in the experiments are all decoder-only transformer \cite{radford2018improving} models generally called \enquote{generative LLMs}. We used a range of instruction-tuned LLMs of different parameter sizes (from 27B for Gemma to at least 70B for Llama3, while proprietary models are expected to be much larger) to cover various models, from open weights to proprietary, general purpose or, in the case of Jais, ones that specifically target the English-Arabic pair \cite{sengupta2023jais}. 

In this paper, \enquote{Llama} refers \texttt{Llama 3.3 70B-Instruct} \citep{dubey2024Llama}, \enquote{GPT-4o} \citep{gpt4o2024gpt} is \texttt{gpt-4o-2024-08-06},\footnote{\url{https://arxiv.org/pdf/2203.02155}} \enquote{Claude} to \citet{claudeanthropic}'s 3.5 Sonnet,\footnote{\texttt{anthropic.claude-3-5-sonnet-20240620-v1:0}} \enquote{Gemma} to \texttt{gemma-2-27b-it} \cite{team2024gemma}, and, also from Google, \enquote{Gemini} to the latest Gemini 1.5 Pro version \cite{team2024gemini}.\footnote{December 2024 release.}
\enquote{Mistral} is \texttt{Mistral Large 24.11} from the eponymous company and, finally, \enquote{Jais} refers to \texttt{jais-family-30b-16k-chat} \cite{sengupta2023jais}.

\subsection{Prompts}
\label{subsec:prompts}
In order to achieve the best translation results, we built on \citet{he-2024-prompting}'s findings by assigning the role of a professional translator to our LLM. This approach outperformed both simpler prompts and those with excessive context.
Furthermore, for each of the three main dialects, we used three prompt strategies: no-shot, one-shot, and two-shot, all written in English. These prompts were the same across regions, except for the specific mention of each dialect in the corresponding prompts. The examples used in the one- and two-shot configurations are not part of the evaluated set, and are from the Algerian and Lebanese dialects. 
We further refined and duplicated these prompt variations to cover two target languages: one set asked for translation into English and the other into MSA. Finally, all prompts were translated into Arabic by a native Arabic speaker who is also a professional translator.
In total, we ran experiments with 36 unique prompts (3 regions * 3 strategies * 2 target languages * 2 prompt languages), or 18 per target language, which we used on all models. The prompts in English are available in Appendix~\ref{app:prompt}, while their equivalents in Arabic can be found  \href{https://docs.google.com/document/d/1f3qq_S9Qbi9QOCgQtHYtiP0CbGp7V4LVcl6k3ZJmILg/edit?usp=sharing}{here}.

\subsection{MT Evaluation}
\label{subsec:eval}
We used automatic metrics and evaluated the potential of using LLM-as-a-judge for direct assessment evaluation.

\subsubsection{Automatic Evaluation}
\label{subsec:metrics}
We used several metrics to quantify the quality of the generated translations. On the more classical side we use BLEU \citep{papineni-etal-2002-bleu}, chrF \citep{popovic-2015-chrF} and TER \citep{snover-etal-2006-study}. All scores were calculated
using SacreBLEU \citep{post-2018-call}.\footnote{The relevant signatures are provided in Appendix~\ref{app:signatures}.}
In order to avoid the usual pitfalls of word- and character-based metrics, especially since we were studying dialects without formal orthography, we further investigated the quality of the translations using techniques based on sub-word embeddings: BERTScore \citep{zhang2019bertscore} and two versions of COMET: COMET-22 \citep[][\texttt{Unbabel/wmt22-comet-da}]{rei-etal-2022-comet} and its reference-free version CometKiwi \citep[][\texttt{Unbabel/wmt22-cometkiwi-da}]{rei-etal-2022-cometkiwi}. The latter three methods help alleviate two limitations of our work: the fact that only one reference translation is available for each sentence and the extremely short length of certain sentences.

\subsubsection{Human Evaluation}
Since no Arabizi-specific metric or resource exists for our dialect selection, we assessed whether LLMs in an \enquote{LLM-as-a-judge} setting \citep{zheng2023judging} can be used to mimic human evaluation to reduce the reliance on hard-to-source users of Arabizi.

For human evaluation, we adopted the direct assessment method, which evaluates translations based on a Likert scale ranging from 1 to 5 (higher is better) according to two key criteria: fluency and adequacy (See Appendix~\ref{app:adequacy_fluency}).
Due to time and human resource constraints, we did not manually annotate all translations. We instead sampled a random machine translation for both target languages and for each source sentence of our dataset. These machine translation outputs were sampled across all our variables, i.e.\ models, prompt languages, and prompt strategies. The resulting set, consisting of 671 segments (268 for Lebanon, 159 for Egypt and 244 for Algeria, see Table~\ref{tab:merged_summary}) for each target language, was then manually rated by two native speakers of Arabic who are professional translators, one of them being the first author of this study and the original translator of the dataset. We then calculated \citet{cohen1960coefficient} $\kappa$ to measure their agreement in terms of fluency and accuracy (see  Appendix~\ref{app:correlations} for results). 
\citet{cohen1960coefficient} $\kappa$ results indicate moderate agreement for adequacy and lower agreement for fluency, with some variations across language pairs. 

The set, which not only consisted of the reference and machine translation but also of the source sentence, was then iteratively fed into GPT-4o in an \enquote{LLM-as-a-judge} setting \citep{zheng2023judging}, with a prompt in English tasking the LLM to follow the human annotation guidelines.\footnote{The prompt is shared in Appendix~\ref{sec:prompt-judge}.} The LLM showed strong correlation with both human annotators, with \citet{spearman1904}'s $\rho$s comprised in the range from $0.457$ (annotator 2, fluency, Egypt to EN) to $0.844$ (annotator 2, adequacy, Egypt to AR). These results indicate that an LLM can be used as an easy way to gauge translation quality during model development. The different correlations as well as all the data for direct assessment can be found in Appendix~\ref{app:correlations}. 

\section{Results}
\subsection{Qualitative Error Analysis}
Due to space constraints, this section will only provide some examples of the main errors. Refer to the \href{https://docs.google.com/spreadsheets/d/1_TrWkwJaB2pNJb8Di1dWFOJ_MgUrMKc8bHCuZkjLQJw/edit?usp=sharing}{table} for a more detailed overview of the main errors.
Most models tend to mistranslate, especially when figurative language is used.
A larger issue lies with Llama3 which tends to output words in another script when translating to MSA. An obvious example is the translation of the segment \textit{Almatar da} (\<هذا المطار>, \enquote{this airport}) as \<المطر> \selectlanguage{russian}да\selectlanguage{english} -- transforming the \enquote{da} in Arabic to Cyrillic. The problem is not limited to Cyrillic, as characters in Latin and Chinese scripts can also be found in the output.
Another type of failure specific to a model is Jais' re-occuring hallucinations. The model often associates feelings of anger to an otherwise neutral message, leading to translations that are irrelevant and contain violent information.

\subsection{Quantitative Analysis}



 See Appendix~\ref{tab:translation_performance} for the complete set of results. 
\subsubsection{Effect of Prompting Techniques}
On one hand, one-shot prompting for translations into English increased BLEU scores across all models. For example, in GPT-4o, the BLEU score improved from 17.386 for no-shot to 20.158 for one-shot, a 16\% increase.    
Two-shot prompting, however, provided only a marginal gain or even slight variation. For instance, in GPT-4o, the BLEU score slightly dropped from 20.158 for one-shot to 19.771 for two-shot.
On the other hand, the improvements to translations into Arabic were less pronounced, suggesting that few-shot prompting is less effective. In GPT-4o, BLEU increased from 8.395 for no-shot to 10.099 for one-shot, a 20\% increase, but the shift from one-shot to two-shot (10.150) was minimal.  
Similarly, in the case of Claude-3, the BLEU score improved from 2.982 for no-shot to 4.009 for one-shot, a 34\% increase, but two-shot promting (4.016) provided almost no additional benefit.

\subsubsection{Effect of Target Language}

English translations consistently outperformed Arabic translations across all metrics, indicating that models handle English more effectively. For instance, GPT-4o achieved higher BLEU scores in English (17.39 to 20.16) than in Arabic (8.40 to 10.15), with chrF scores following a similar trend (43.08 to 45.50 for English vs. 36.64 to 38.09 for Arabic). TER also confirmed that English translations required fewer edits, with scores of 70.29 for one-shot compared to 78.70 for Arabic. Other models, such as Claude-3 and Llama-3, exhibited similar disparities, with English BLEU scores nearly doubling those of Arabic. Both COMET metrics and BERTScore further highlighted this gap, although BERTScore pointed to different alignment characteristics between languages. While GPT-4o and Gemini were the strongest models for Arabic, their scores still lagged behind their English performance, reinforcing the overall trend of English translations being more accurate and consistent.

\subsubsection{Effect of Source Dialect}
The evaluation of translation performance across different dialects revealed notable variations in quality, as measured by the different translation metrics (cf Appendix~\ref{tab:translation_scores}). The Egyptian dialect demonstrated the highest translation quality, with an average BLEU score of 9.65 and a chrF score of 34.64, indicating the highest word- and character-level accuracy. Additionally, Egyptian achieved a BERTScore of 0.37 and a COMET score of 0.67, suggesting higher semantic similarity to reference translations. The Lebanese dialect followed with a BLEU score of 7.52 and a chrF score of 26.59, with a comparable COMET Kiwi score of 0.48 but a slightly lower COMET score of 0.65. The Algerian dialect ranked third, with a significantly lower BLEU score of 4.24 and a chrF score of 23.21, along with the lowest BERTScore of 0.33 and COMET score of 0.63.

The disparity in translation quality among the dialects could be explained by linguistic, sociocultural, and technological factors. Egyptian Arabic, the most widely spoken and documented dialect, aligns closely with MSA and is predominant in the media, ensuring better representation in training datasets. 
By contrast, Algerian Arabic’s heavy code-switching (cf Table~\ref{tab:merged_summary}) with Berber, French, and Spanish, along with figurative word meanings, make translation more challenging. Its lack of representation in digital corpora further limits LLMs training, resulting in poorer translation performance.

\subsubsection{Effect of Prompt Language}
As seen in a prior article \citep{zhang2023promptinglargelanguagemodel}, our results confirm that prompting in English generally yields better results across all models.

\subsection{Metrics Correlation}
Because traditional metrics such as BLEU and chrF quantify n‐gram overlap with the reference, thereby rewarding surface‐level similarity and penalizing deviations, they tend to produce correlated scores and inversely correlate with TER. 

Meanwhile, embedding‐based metrics such as BERT Score and COMET rely on learned contextual representations to gauge semantic similarity, thus capturing deeper nuances in meaning and tolerating surface‐level variations, which often leads them to yield patterns that are distinct from n‐gram‐focused measures.

Across the different combinations, BLEU and chrF scores typically fluctuated in parallel. However, certain model-prompt settings revealed inconsistencies, where BLEU increased while BERT Score or COMET remained unchanged or declined, indicating improved n-gram overlap but not necessarily better semantic accuracy or fluency. Despite these inconsistencies, higher BLEU generally correlated with good embedding-based metrics scores.

\section{Conclusion and Discussion}

Models struggle significantly with Arabizi. GPT-4o is the best-performing translation model, followed by Gemini. Mistral Large and Gemma perform moderately well, while Llama 3 and Jais are the weakest models (see Appendix~\ref{app:model_ranking}). Interestingly, Gemma performed surprisingly well in translation tasks despite being a 27B parameter model. Its results, particularly in English, were competitive with larger models, suggesting that model size is not the only determinant of translation quality—architectural optimizations and training data also play a crucial role.

For model prompting, few-shot approaches improved performance but was more effective for English than for Arabic. English prompts worked better overall and in all prompting scenarios, though the difference was much less stark for GPT-4o and Gemini and, to a lesser extent, for Gemma. 

Despite a large variation in average segment length between different dialects, no clear pattern emerged in terms of automatic scores. This hints that translation quality does not directly depend on segment length. 

The LLM-as-a-Judge scenario aligned with expert human raters, making it a relevant tool in this setting. This study further shows that while far from perfect, using \enquote{out-of-the-box} LLMs to translate Arabizi is a viable solution for gisting, especially when combined with an LLM-as-a-judge.

\section{Limitations}

This study has several limitations. First, the dataset does not fully capture the diversity of Arabizi usage across different regions and social contexts. Second, it relies on translators who are non-native speakers of English. Third, the variety of text lengths may affect performance, as shorter or longer texts might yield varying results. 
Furthermore, no Arabizi-specific evaluation metric was used, which can affect the accuracy of the assessments. Lastly, the study was constrained by a relatively small corpus, which may limit the applicability of its findings.

\section{Acknowledgements}
We would like to thank the volunteers who donated their conversations, the translator, the reviewers for their helpful comments and suggestions. Finally, our thanks go to Dr. Thien for her proofreading assistance.

\section{CO$_2$ Emission Related to Experiments}
It is difficult to estimate the energy usage of models that were run in an \enquote{inference-as-a-service} setting, especially when the details of such models are proprietary. Using the tool provided by \citet{lannelongue2021green}\footnote{\url{https://calculator.green-algorithms.org/ai}} and basing our calcuations on model sizes of around 400B parameters for the proprietary models, we estimate that the energy usage of our experiments amounted to $6.99$ kWh in a US datacenter, which corresponds to a carbon footprint of $2.97$ kgCO$_2$e.

\bibliography{anthology,mtsummit25}

\clearpage
\appendix

\onecolumn
\section{Prompts in English for all Dialects, Target Languages, and Prompt Strategies}
\label{app:prompt}
\enquote{ALG} stands for Algeria, \enquote{EG} for Egypt, and \enquote{LB} for Lebanon. 
For the experiments with a prompt in Arabic, all prompts were translated into Modern Standard Arabic by the first author of the study, who is a native speaker of Arabic and a professional translator.
\lstset{language=JSON, escapeinside=``}
\begin{lstlisting}
{
    "ALG_AR": {
        "no-shot": "You are a professional Arabic translator with years of experience translating spoken language from various Arabic dialects. Translate the following text from the Algerian dialect to Modern Standard Arabic.",
        
        "one-shot": "You are a professional Arabic translator with years of experience translating spoken language from various Arabic dialects.
        
            Source text: "Ma 3am efham chu 3am te7ke"
            Target text in Arabic: "`\<لا أفهم ما تقوله>`"
        
        Based on the example above, translate the following text from the Algerian dialect to Modern Standard Arabic.",
        
        "two-shot": "You are a professional Arabic translator with years of experience translating spoken language from various Arabic dialects.
        
            Source text: "Ma 3am efham chu 3am te7ke"
            Target text in Arabic: "`\<لا أفهم ما تقوله>`"
        
            Source text: "M t7kilich 7yetk kho"
            Target text in Arabic: "`\<لا تقص عليّ قصّة حياتك أخي>`"
        
        
        Based on the examples above, translate the following text from the Algerian dialect to Modern Standard Arabic."
    },
    "ALG_EN": {
        "no-shot": "You are a professional Arabic translator with years of experience translating spoken language from various Arabic dialects. Translate the following text from the Algerian dialect to English.",
        
        "one-shot": "You are a professional Arabic translator with years of experience translating spoken language from various Arabic dialects.
    
            Source text: "Ma 3am efham chu 3am te7ke"
            Target text in English: "I don't understand what you're saying."
                
        Based on the example above, translate the following text from the Algerian dialect to English.",
        
        "two-shot": "You are a professional Arabic translator with years of experience translating spoken language from various Arabic dialects.
        
            Source text: "Ma 3am efham chu 3am te7ke"
            Target text in English: "I don't understand what you're saying."
        
            Source text: "M t7kilich 7yetk kho"
            Target text in English: "Don't tell me your life story, bro"
            
        Based on the examples above, translate the following text from the Algerian dialect to English."
        },
    "EG_AR": {
        "no-shot": "You are a professional Arabic translator with years of experience translating spoken language from various Arabic dialects. Translate the following text from the Egyptian dialect to Modern Standard Arabic.",
        
        "one-shot": "You are a professional Arabic translator with years of experience translating spoken language from various Arabic dialects.

            Source text: "Ma 3am efham chu 3am te7ke"
            Target text in Arabic: "`\<لا أفهم ما تقوله>`"
                
        Based on the example above, translate the following text from the Egyptian dialect to Modern Standard Arabic.",
        
        "two-shot": "You are a professional Arabic translator with years of experience translating spoken language from various Arabic dialects.
        
            Source text: "Ma 3am efham chu 3am te7ke"
            Target text in Arabic: "`\<لا أفهم ما تقوله>`"
            
            Source text: "M t7kilich 7yetk kho"
            Target text in Arabic: "`\<لا تقص عليّ قصّة حياتك أخي>`"
        
        Based on the examples above, translate the following text from the Egyptian dialect to Modern Standard Arabic."
    },
    "EG_EN": {
        "no-shot": "You are a professional Arabic translator with years of experience translating spoken language from various Arabic dialects. Translate the following text from the Egyptian dialect to English.",
        
        "one-shot": "You are a professional Arabic translator with years of experience translating spoken language from various Arabic dialects.

            Source text: "Ma 3am efham chu 3am te7ke"
            Target text in English: "I don't understand what you're saying."

        Based on the example above, translate the following text from the Egyptian dialect to English.",

        "two-shot": "You are a professional Arabic translator with years of experience translating spoken language from various Arabic dialects.

            Source text: "Ma 3am efham chu 3am te7ke"
            Target text in English: "I don't understand what you're saying."
    
            Source text: "M t7kilich 7yetk kho"
            Target text in English: "Don't tell me your life story, bro"

        Based on the examples above, translate the following text from the Egyptian dialect to English."
        },
    "LB_AR": {
        "no-shot": "You are a professional Arabic translator with years of experience translating spoken language from various Arabic dialects. Translate the following text from the Lebanese dialect to Modern Standard Arabic.",
        
        "one-shot": "You are a professional Arabic translator with years of experience translating spoken language from various Arabic dialects.

            Source text: "Ma 3am efham chu 3am te7ke"
            Target text in Arabic: "`\<لا أفهم ما تقوله>`"
            

        Based on the example above, translate the following text from the Lebanese dialect to Modern Standard Arabic.",
        
        "two-shot": "You are a professional Arabic translator with years of experience translating spoken language from various Arabic dialects.

            Source text: "Ma 3am efham chu 3am te7ke"
            Target text in Arabic: "`\<لا أفهم ما تقوله>`"
            
            Source text: "M t7kilich 7yetk kho"
            Target text in Arabic: "`\<لا تقص عليّ قصّة حياتك أخي>`"
            
        Based on the examples above, translate the following text from the Lebanese dialect to Modern Standard Arabic."
        },
        "LB_EN": {
        "no-shot": "You are a professional Arabic translator with years of experience translating spoken language from various Arabic dialects. Translate the following text from the Lebanese dialect to English.",
        
        "one-shot": "You are a professional Arabic translator with years of experience translating spoken language from various Arabic dialects.
    
            Source text: "Ma 3am efham chu 3am te7ke"
            Target text in English: "I don't understand what you're saying."
            
        Based on the example above, translate the following text from the Lebanese dialect to English.",
        
        "two-shot": "You are a professional Arabic translator with years of experience translating spoken language from various Arabic dialects.
        
            Source text: "Ma 3am efham chu 3am te7ke"
            Target text in English: "I don't understand what you're saying."
            
            Source text: "M t7kilich 7yetk kho"
            Target text in English: "Don't tell me your life story, bro"
        
        Based on the examples above, translate the following text from the Lebanese dialect to English."
    }
}
\end{lstlisting}

\section{Adequacy and Fluency}
\label{app:adequacy_fluency}
\begin{table*}[h]
    \centering
    \begin{tabular}{lll}
        \toprule
        \textbf{Score} & \textbf{Adequacy} & \textbf{Fluency} \\
        \midrule
        5 & All Meaning & Flawless Language \\
        4 & Most Meaning & Good Language  \\
        3 & Much Meaning & Non-native Language \\
        2 & Little Meaning & Disfluent Language \\
        1 & None & Incomprehensible Language\\
        \bottomrule
    \end{tabular}
    \caption{Adequacy and Fluency Evaluation Scale \cite{km2006}}
    \label{tab:adequacy_fluency}    
\end{table*}

\section{LLM-as-a-judge}
\label{sec:prompt-judge}

The system prompt was the following: 
\begin{displayquote}
You are a professional translator, expert in Arabic, English, and Arabic dialects. Your role here is to evaluate the quality of a translation using two dimensions: `Adequacy' (scale of 1 to 5, higher is better) and `Fluency' (scale of 1 to 5, higher is better). You will be given a source text in Arabic dialect, a reference translation into \texttt{\{target\_lang\}}, and a machine translation. Return in this format, and NOTHING ELSE:
\begin{verbatim}
Adequacy:[your_score]
Fluency:[your_score]
\end{verbatim}
I trust and count on you.
\end{displayquote}

\hspace{-1em}The prompt was the following: 
\begin{displayquote}
Source from \texttt{\{country\}}: \texttt{\{source\}} \\
Reference translation: \texttt{\{ref\}} \\
Machine translation: \texttt{\{hyp\}} \\
Give scores from 1 to 5 for both Adequacy and Fluency using the template:
\begin{verbatim}
Adequacy:[your_score]
Fluency:[your_score]
\end{verbatim}
Return nothing else.
\end{displayquote}

\section{Metrics Signatures}
\label{app:signatures}
\begin{flushleft}
\texttt{BLEU: nrefs:1|case:mixed|eff:no|tok:13a|smooth:exp|version:2.5.1}\\
\texttt{chrF: nrefs:1|case:mixed|eff:yes|nc:6|nw:0|space:no|version:2.5.1}\\
\texttt{TER: nrefs:1|case:lc|tok:tercom|norm:no|punct:yes|}\\
\texttt{\hspace*{2.4em}asian:no|version:2.5.1}
\end{flushleft}

\section{Human-LLM-as-a-judge Correlation and Direct Assessment}
\label{app:correlations}
\begin{table*}[h!]
\centering
    \begin{tabular}{lrrrr}
    \toprule
        \multirow{2}{*}{\textbf{Country}} & \multicolumn{2}{c}{\textbf{Fluency}} & \multicolumn{2}{c}{\textbf{Adequacy}} \\
         & \textbf{LLM-Rater 1} & \textbf{LLM-Rater 2} & \textbf{LLM-Rater 1} & \textbf{LLM-Rater 2} \\
        \midrule
            Lebanon - EN   & 0.602 & 0.495 & 0.653 & 0.824 \\
            Lebanon - AR   & 0.685 & 0.601 & 0.820 & 0.795 \\
            \midrule
            Egypt - EN     & 0.631 & 0.457 & 0.667 & 0.781 \\
            Egypt - AR     & 0.637 & 0.611 & 0.677 & 0.844 \\
            \midrule
            Algeria - EN   & 0.642 & 0.536 & 0.683 & 0.800 \\
            Algeria - AR   & 0.678 & 0.485 & 0.760 & 0.770 \\
            \bottomrule
    \end{tabular}
    \caption{Correlation Scores (\citet{spearman1904}'s $\rho$) Between Human Annotators and LLM-as-a-Judge in Direct Assessment Scores per Country and Target Language.}
\end{table*}

\begin{table*}[h!]
\centering
    \begin{tabular}{lcccc}
    \toprule
    \multirow{2}{*}{\textbf{Country}} & \multicolumn{2}{c}{\textbf{Fluency}} & \multicolumn{2}{c}{\textbf{Adequacy}} \\
     & \textbf{Rater 1} & \textbf{Rater 2} & \textbf{Rater 1} & \textbf{Rater 2} \\
    \midrule
    Lebanon - EN & 2.494 & 3.822 & 2.203 & 2.431 \\
    Lebanon - AR & 2.782 & 3.430 & 2.362 & 2.662 \\
    \midrule
    Egypt - EN & 2.560 & 3.340 & 2.082 & 2.679 \\
    Egypt - AR & 2.956 & 3.538 & 2.497 & 2.887 \\
    \midrule
    Algeria - EN & 2.534 & 3.773 & 1.853 & 2.315 \\
    Algeria - AR & 2.721 & 3.500 & 2.225 & 2.335 \\
    \bottomrule
    \end{tabular}
    \caption{Average Direct Assessment Scores (1-5) for Both Human Raters, per Country and Target Language.}
\end{table*}

\clearpage
\section{Automatic Evaluation Results for all Models}
\label{tab:translation_performance}
Due to space constraints, we are presenting the results averaged over the two prompt languages. A more comprehensive overview of the results is available in the accompanying \href{https://docs.google.com/spreadsheets/d/112HUEHM3SIDgarizmqMBOD7OjNmzGJcSUOtgt_5SuIk/edit?usp=sharing}{spreadsheet}.

\begin{table*}[h!]
    \centering
    \small
    \begin{tabular}{l l l r r r r r r}
        \toprule
        \textbf{Model} & \textbf{TL} & \textbf{Prompt Tech} & \textbf{BLEU} & \textbf{chrF} & \textbf{TER} & \textbf{BERT} & \textbf{KIWI} & \textbf{COMET} \\
        \midrule
        GPT-4o & EN & no-shot & 17.386 & 43.081 & 77.038 & 0.478 & 0.434 & 0.733 \\
        GPT-4o & EN & one-shot & \textbf{20.158} & 45.232 & \textbf{70.287} & \textbf{0.529} & \textbf{0.439} & \textbf{0.757} \\
        GPT-4o & EN & two-shot & 19.771 & \textbf{45.496} & 70.294 & 0.521 & \textbf{0.439} & 0.755 \\
        \midrule
        GPT-4o & AR & no-shot & 8.395 & 36.637 & 86.501 & 0.558 & 0.422 & 0.757 \\
        GPT-4o & AR & one-shot & 10.099 & \textbf{38.221} & \textbf{78.701} & \textbf{0.586} & \textbf{0.423} & \textbf{0.776} \\
        GPT-4o & AR & two-shot & \textbf{10.150} & 38.090 & 79.775 & 0.585 & 0.421 & 0.774 \\
        \midrule
        claude3 & EN & no-shot & 5.603 & 25.565 & 150.074 & \textbf{0.270} & 0.540 & 0.620 \\
        claude3 & EN & one-shot & 8.795 & 30.356 & 97.400 & 0.191 & 0.536 & 0.605 \\
        claude3 & EN & two-shot & \textbf{9.433} & \textbf{32.360} & \textbf{96.325} & 0.225 & \textbf{0.541} & \textbf{0.625} \\
        \midrule
        claude3 & AR & no-shot & 2.982 & 22.957 & 122.794 & 0.367 & 0.428 & 0.657 \\
        claude3 & AR & one-shot & 4.009 & 28.169 & \textbf{97.126} & 0.420 & \textbf{0.431} & \textbf{0.686} \\
        claude3 & AR & two-shot & \textbf{4.016} & \textbf{28.473} & 98.488 & \textbf{0.425} & 0.427 & \textbf{0.686} \\
        \midrule
        Llama3 & EN & no-shot & 6.972 & 27.982 & 107.160 & \textbf{0.293} & \textbf{0.526} & \textbf{0.620} \\
        Llama3 & EN & one-shot & \textbf{8.234} & \textbf{28.510} & 101.095 & 0.290 & 0.518 & 0.618 \\
        Llama3 & EN & two-shot & 7.709 & 27.988 & \textbf{99.623} & 0.274 & 0.521 & 0.613 \\
        \midrule
        Llama3 & AR & no-shot & 2.196 & 17.748 & 160.334 & 0.243 & 0.412 & 0.587 \\
        Llama3 & AR & one-shot & \textbf{2.862} & \textbf{20.763} & \textbf{118.675} & \textbf{0.343} & \textbf{0.425} & \textbf{0.623} \\
        Llama3 & AR & two-shot & 1.139 & 17.728 & 218.553 & 0.286 & \textbf{0.425} & 0.610 \\
        \midrule
        gemma2 & EN & no-shot & 8.523 & 27.979 & 89.761 & 0.330 & \textbf{0.548} & 0.634 \\
        gemma2 & EN & one-shot & 9.866 & 28.977 & 88.589 & \textbf{0.337} & 0.543 & 0.639 \\
        gemma2 & EN & two-shot & \textbf{9.925} & \textbf{29.258} & \textbf{87.319} & 0.333 & 0.545 & \textbf{0.640} \\
        \midrule
        gemma2 & AR & no-shot & 3.583 & 23.098 & 99.113 & 0.358 & \textbf{0.429} & \textbf{0.644} \\
        gemma2 & AR & one-shot & \textbf{4.070} & 24.175 & \textbf{94.968} & 0.388 & 0.424 & 0.660 \\
        gemma2 & AR & two-shot & 3.959 & \textbf{24.400} & 96.123 & \textbf{0.389} & 0.426 & \textbf{0.664} \\
        \midrule
        mistrallarge & EN & no-shot & 7.919 & \textbf{29.002} & 101.704 & \textbf{0.285} & \textbf{0.524} & \textbf{0.627} \\
        mistrallarge & EN & one-shot & \textbf{9.409} & 28.445 & 91.376 & 0.263 & 0.518 & 0.610 \\
        mistrallarge & EN & two-shot & 9.259 & 28.347 & \textbf{91.020} & 0.268 & 0.521 & 0.612 \\
        \midrule
        mistrallarge & AR & no-shot & 4.019 & \textbf{25.413} & 103.042 & \textbf{0.434} & \textbf{0.493} & \textbf{0.673} \\
        mistrallarge & AR & one-shot & 4.230 & 24.663 & 95.517 & 0.428 & 0.482 & 0.672 \\
        mistrallarge & AR & two-shot & \textbf{4.372} & 25.162 & \textbf{94.433} & 0.426 & 0.481 & 0.669 \\
        \midrule
        jais & EN & no-shot & 1.518 & 15.273 & 344.316 & \textbf{0.065} & 0.470 & 0.501 \\
        jais & EN & one-shot & \textbf{2.157} & 15.667 & \textbf{160.450} & 0.016 & \textbf{0.511} & 0.507 \\
        jais & EN & two-shot & 1.735 & \textbf{15.958} & 192.409 & 0.042 & 0.499 & \textbf{0.508} \\
        \midrule
        jais & AR & no-shot & 0.750 & \textbf{14.486} & 208.837 & 0.241 & \textbf{0.420} & \textbf{0.583} \\
        jais & AR & one-shot & \textbf{0.847} & 14.120 & \textbf{196.811} & 0.258 & 0.418 & 0.578 \\
        jais & AR & two-shot & 0.704 & 14.418 & 202.979 & \textbf{0.270} & 0.418 & 0.579 \\
        \midrule
        gemini & EN & no-shot & 11.317 & 36.950 & 94.146 & 0.379 & 0.493 & 0.672 \\
        gemini & EN & one-shot & 16.119 & 41.023 & 78.911 & 0.451 & 0.493 & 0.713 \\
        gemini & EN & two-shot & \textbf{16.187} & \textbf{41.182} & \textbf{77.605} & \textbf{0.455} & \textbf{0.495} & \textbf{0.720} \\
        \midrule
        gemini & AR & no-shot & 5.174 & 31.319 & 90.174 & 0.502 & \textbf{0.470} & 0.729 \\
        gemini & AR & one-shot & 6.636 & 33.513 & 84.945 & 0.536 & 0.468 & 0.752 \\
        gemini & AR & two-shot & \textbf{7.585} & \textbf{33.987} & \textbf{84.428} & \textbf{0.541} & \textbf{0.470} & \textbf{0.753} \\
        \bottomrule
    \end{tabular}
    \caption{Automatic Evaluation Results for all Models, Averaged over Prompt Languages. TL = Target Language, BERT = BERTScore, KIWI = wmt22-cometkiwi-da, COMET = wmt22-comet-da. Best scores for every model, target language, and prompt strategy are indicated in bold.}
\end{table*}
\clearpage


\section{Average Metric Scores for Dialects}
\begin{table*}[h!]
    \centering
    \begin{tabular}{lcccccc}
        \toprule
        \textbf{Country} & \textbf{BLEU} & \textbf{chrF} & \textbf{TER} & \textbf{BERTScore} & \textbf{KIWI} & \textbf{COMET} \\
        \midrule
        Lebanon & 7.5212 & 26.5935 & 123.5180 & 0.3604 & 0.4799 & 0.6547 \\
        Egypt   & 9.6466 & 34.6404 & 102.8187 & 0.3699 & 0.4749 & 0.6749 \\
        Algeria & 4.2445 & 23.2068 & 122.1789 & 0.3322 & 0.4642 & 0.6303 \\
        \bottomrule
    \end{tabular}
    \caption{Translation Quality Scores for the Arabic Dialects}
    \label{tab:translation_scores}
\end{table*}

\section{Model Ranking}
\label{app:model_ranking}
\begin{table*}[h!]
\centering
\begin{tabular}{lrrrrrr}
\toprule
\textbf{Model} &       \textbf{BLEU} &       \textbf{chrF} &         \textbf{TER} &  \textbf{BERTScore} &  \textbf{KIWI} &  \textbf{COMET} \\
\midrule
GPT-4o        &  14.326 &  41.126 &   77.099 &    0.542 &     0.429 &   0.758 \\
gemini       &  10.503 &  36.329 &   85.034 &    0.477 &     0.481 &   0.723 \\
gemma2       &   6.654 &  26.314 &   92.645 &    0.355 &     0.485 &   0.646 \\
mistrallarge &   6.534 &  26.838 &   96.181 &    0.350 &     0.503 &   0.643 \\
claude3      &   5.806 &  27.980 &  110.368 &    0.316 &     0.483 &   0.646 \\
Llama3       &   4.851 &  23.453 &  134.240 &    0.287 &     0.471 &   0.611 \\
jais         &   1.285 &  14.986 &  217.633 &    0.148 &     0.456 &   0.542 \\
\bottomrule
\end{tabular}
\caption{Ranking of Translation Models from Best to Worst Based on Average Automatic Metric Scores}
\end{table*}

\end{document}